\documentclass{article}

\usepackage[preprint]{neurips_2025}

\usepackage[utf8]{inputenc} % allow utf-8 input
\usepackage[T1]{fontenc}    % use 8-bit T1 fonts
\usepackage{hyperref}       % hyperlinks
\usepackage{url}            % simple URL typesetting
\usepackage{booktabs}       % professional-quality tables
\usepackage{amsfonts}       % blackboard math symbols
\usepackage{nicefrac}       % compact symbols for 1/2, etc.
\usepackage{microtype}      % microtypography
\usepackage{xcolor}         % colors

\usepackage{graphicx}
\usepackage{subfigure}
\usepackage{amsmath}
\usepackage{multirow}

\title{GET: Goal-directed Exploration and Targeting for Large-Scale Unknown Environments}

\author{%
  Lanxiang Zheng \\
  School of Computer Science and Engineering\\
  Sun Yat-Sen University\\
  Guangzhou, 511400 \\
  \texttt{zhenglx35@mail2.sysu.edu.cn} \\
  \And
  Ruidong Mei \\
  School of Systems Science and Engineering\\
  Sun Yat-Sen University\\
  Guangzhou, 511400 \\
  \texttt{meird@mail2.sysu.edu.cn}
  \And
  Mingxin Wei \\
  School of Artificial Intelligence\\
  Sun Yat-Sen University\\
  Zhuhai, 519082 \\
  \texttt{weimx3@mail2.sysu.edu.cn}
  \And
  Hao Ren \\
  School of Computer Science and Engineering\\
  Sun Yat-Sen University\\
  Guangzhou, 511400 \\
  \texttt{renh55@mail2.sysu.edu.cn}
  \And
  Hui Cheng \\
  School of Computer Science and Engineering\\
  Sun Yat-Sen University\\
  Guangzhou, 511400 \\
  \texttt{chengh9@mail.sysu.edu.cn}
}

\begin{document}

\maketitle

\begin{abstract}
  Object search in large-scale, unstructured environments remains a fundamental challenge in robotics, particularly in dynamic or expansive settings such as outdoor autonomous exploration. This task requires robust spatial reasoning and the ability to leverage prior experiences. While Large Language Models (LLMs) offer strong semantic capabilities, their application in embodied contexts is limited by a grounding gap in spatial reasoning and insufficient mechanisms for memory integration and decision consistency.To address these challenges, we propose GET (Goal-directed Exploration and Targeting), a framework that enhances object search by combining LLM-based reasoning with experience-guided exploration. At its core is DoUT (Diagram of Unified Thought), a reasoning module that facilitates real-time decision-making through a role-based feedback loop, integrating task-specific criteria and external memory. For repeated tasks, GET maintains a probabilistic task map based on a Gaussian Mixture Model, allowing for continual updates to object-location priors as environments evolve.Experiments conducted in real-world, large-scale environments demonstrate that GET improves search efficiency and robustness across multiple LLMs and task settings, significantly outperforming heuristic and LLM-only baselines. These results suggest that structured LLM integration provides a scalable and generalizable approach to embodied decision-making in complex environments.
\end{abstract}

\section{Introduction}
Object searching is a fundamental task in mobile robotics, with broad applications in service robotics\cite{jaboob2024artificial}, search-and-rescue\cite{martinez2021search}, and autonomous exploration\cite{garaffa2021reinforcement}. Achieving efficient and reliable object search in complex environments requires robots to perceive their surroundings, reason about spatial layouts, and make sequential decisions in real time\cite{sun2024survey}.

Despite recent progress, object search remains a challenging problem, particularly in large-scale, unstructured environments. Traditional approaches typically rely on hand-crafted search strategies or heuristic exploration rules\cite{tsuru2021online, luo2024star, zhang2024hierarchical}, which often fail to generalize across diverse tasks and environments. Recent advances in deep learning and reinforcement learning have led to notable improvements by enabling data-driven policies\cite{kulkarni2016hierarchical, ye2021efficient, khanna2024goat}. However, these methods still face generalization issues due to their reliance on exhaustive and task-specific training data, as well as their limited ability to reason about spatial semantics or incorporate historical experiences.

To address these shortcomings, Large Language Models (LLMs) have emerged as a promising alternative, offering strong contextual understanding and high-level reasoning capabilities\cite{tsuru2021online, luo2024star, zhang2024hierarchical}. By analyzing environmental descriptions and inferring semantic cues, LLMs can guide robot exploration with more adaptive and informed strategies. However, their application to 3D object search remains limited due to several key challenges: weak spatial perception\cite{brohan2023can}, dependence on extensive prompt tuning or fine-tuning\cite{li2022pre}, and difficulty in maintaining robust and consistent performance in dynamic or unstructured environments. Moreover, LLMs lack effective mechanisms for recording or retrieving task-specific historical knowledge, making it challenging to leverage accumulated past experiences\cite{gao2023retrieval, das2024larimar}. These limitations become particularly pronounced in large-scale or outdoor settings, where the spatial correspondence between objects and their surroundings becomes diffuse, increasing ambiguity and complicating the generalization of search strategies across tasks and environments.

To address these limitations, we introduce a novel framework called GET (Goal-directed Exploration and Targeting). GET leverages the contextual reasoning capabilities of Large Language Models (LLMs), enhanced by a new reasoning improvement module, the Diagram of Unified Thought (DoUT), designed specifically to refine decision-making without extensive task-specific pretraining. In the DoUT module, an LLM-generated proposal is iteratively refined through evaluative feedback from an independent evaluator model, promoting self-improvement in real-time decision-making. Additionally, GET incorporates a Gaussian Mixture Model (GMM)-based probabilistic approach to explicitly represent and update historical search experiences. This probabilistic representation incrementally updates the target object's spatial distribution, effectively addressing the limitations of LLMs in utilizing historical data and providing robust guidance for repeated search tasks. The main contributions of this paper are summarized as follows:

\begin{itemize}
\item We propose GET, a novel LLM-based framework for goal-directed exploration and target localization in large-scale environments, integrating semantic reasoning with probabilistic experience modeling.
\item We introduce DoUT, a general-purpose reasoning enhancement module that facilitates iterative and adaptive decision-making in real time, avoiding the need for extensive task-specific annotations.
\item We present a GMM-based probabilistic experience model that allows for explicit memory of prior tasks, supporting incremental updates and the gradual fading of outdated information to accelerate the efficiency of repeated search tasks.
\item We validate the proposed method through real-world experiments, demonstrating improved efficiency and performance in challenging large-scale environmentss.
\end{itemize}

\begin{figure}[tbp]
    \centering
    \includegraphics[width=1.0 \linewidth]{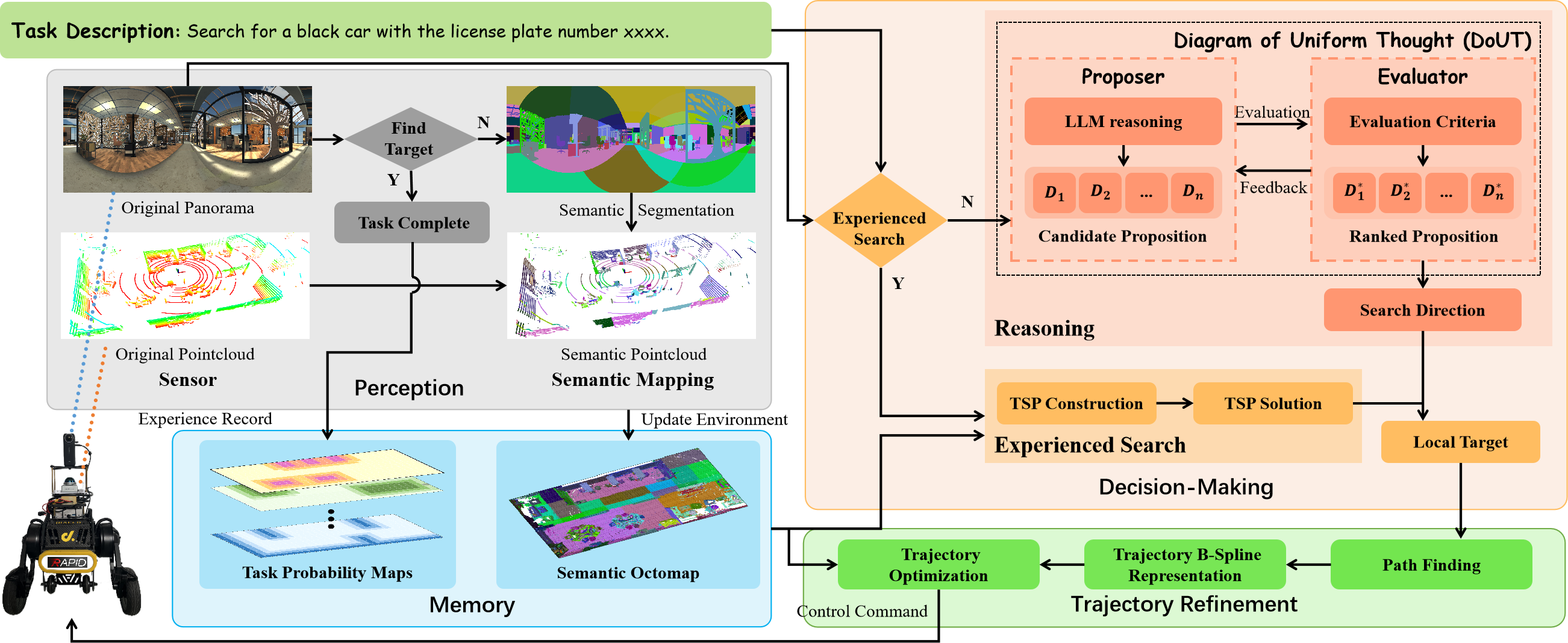}
    \caption{The GET framework consists of several modules: The Perception module converts panoramic images and point clouds into semantic point clouds. The Memory model features a real-time updated semantic octomap and a multi-layered task probability map that record the environment and historical experiences. The Decision-Making module operates in two modes: reasoning search, where the DoUT module infers potential target locations to guide the robot's search process, and experience-based search, where the robot utilizes historical data and the semantic octomap to locate the target and plan search routes. Finally, the Trajectory Refinement module optimizes the robot's path for safe, smooth, and continuous navigation.}
    \label{fig:framework}
\end{figure}

% \begin{itemize}
% \item A novel search framework, GET, is proposed to enhance target object search efficiency in large-scale environments by leveraging the advanced comprehension and reasoning capabilities of LLMs, and explicitly recording task experiences with a GMM-based approach.
% \item A general-purpose reasoning enhancement module, DoUT, is designed to support real-time robot decision-making and guide LLM self-learning through feedback-driven reasoning, thereby eliminating the need for task-specific pretraining.
% \item A GMM-based probabilistic method enables explicit experience recording, allowing incremental updates and the gradual fading of outdated information to accelerate the efficiency of repeated search tasks.
% \item Real-world experiments validate the framework, demonstrating significant improvements in robotic search efficiency and performance in complex, large-scale unknown environments.
% \end{itemize}

\section{Related Work}
\subsection{Target Search Methods}
Numerous effective methods are dedicated to object searching tasks. Traditional approaches often rely on predefined search strategies and Simultaneous Localization and Mapping (SLAM), involving detailed path planning and grid-based mapping with various sensors and predetermined rules \cite{kim2022autonomous,dang2018autonomous,papatheodorou2023finding}. However, their effectiveness heavily depends on sensor accuracy \cite{dang2018autonomous,papatheodorou2023finding} and the quality of environmental models \cite{kulkarni2022autonomous,kim2022autonomous,luo2024star}. While effective, these methods lack flexibility to adapt to diverse search targets and perform poorly in large-scale, unstructured environments. 

Neural network methods \cite{pandey2020trajectory,ye2023real,khanna2024goat,chang2023goat} enhance robotic capabilities through advanced data-driven models. For instance, Pandey and Parhi \cite{pandey2020trajectory} proposed a behavior-based neural network for trajectory planning and target search in mobile robots, improving adaptability in dynamic environments. Ye et al. \cite{ye2023real} integrated CNN and Transformer architectures in UAV systems for robust real-time object detection in high-speed scenarios. End-to-end methods, such as SenseAct-NN in GOAT-Bench \cite{khanna2024goat} and GOAT \cite{chang2023goat}, directly map sensor inputs to navigation paths, improving flexibility but relying heavily on large-scale labeled training data, limiting scalability in data-scarce environments.

Reinforcement learning (RL) methods \cite{rao2021visual,luo2022deep,zhu2017target,fang2022target} also show promise in target search tasks. Kulkarni et al. \cite{kulkarni2016hierarchical} introduced h-DQN, a hierarchical RL framework that efficiently solves long-horizon tasks with sparse rewards. Luo et al. \cite{luo2022deep} extended RL to multi-UAV cooperative target search, optimizing energy efficiency in large environments. Zhu et al. \cite{zhu2017target} combined visual observations and goal representations for indoor navigation. Fang et al. \cite{fang2022target} enhanced navigation by integrating imitation learning with deep RL. Ye et al. \cite{ye2021efficient} employed a hierarchical policy framework to improve RL-based search methods in complex environments. DRL-Searcher \cite{guo2024drl} enhances adaptability and efficiency in target search tasks without relying on the target's motion model. Despite strong performance, these RL methods face challenges related to large-scale training data and generalization.

Strategies integrating vision and semantics provide robots with multimodal perception, accelerating object search tasks \cite{anderson2018vision,moudgil2021soat}. By reasoning about the environment and tasks, robots can predict necessary actions more accurately \cite{li2022envedit,li2023kerm}. The emergence of Large Language Models (LLMs) has further advanced search tasks through improved reasoning capabilities. Zhang et al. \cite{zhang2024hierarchical} modeled relationships between search objects and environments using natural language, creating a framework for quick object localization. Schmalstieg et al. \cite{schmalstieg2023learning} introduced interactive search tasks, guiding agents through complex sequences to find objects. Honerkamp et al. \cite{honerkamp2024language} developed open-vocabulary scene graphs to enhance LLMs' semantic reasoning for object searches. Voxposer \cite{huang2023voxposer} uses LLM-guided 3D voxel-based maps to synthesize robot trajectories, enabling semantic reasoning in complex scenes. Even with obstructed visibility, LLMs can predict target locations, aiding navigation \cite{dorbala2024can}. However, most work focuses on indoor or small-scale environments, with less emphasis on large-scale or outdoor searches, where dynamic elements introduce unpredictable challenges.

\subsection{Search Experiences Storage and Utilization}
Storing historical search experiences is crucial for enhancing efficiency and adaptability in target search tasks. Implicit storage methods capture past experiences without explicit external structures. For instance, Ye et al. \cite{ye2021efficient} utilize hidden states to encode historical observations, enabling robots to recall relevant experiences for improved decision-making. Zhu et al. \cite{zhu2017target} combine visual observations with goal representations, while Fang et al. \cite{fang2022target} and Luo et al. \cite{luo2022deep} extend this concept using reinforcement learning for efficient navigation and multi-UAV target search. Kulkarni et al. \cite{kulkarni2016hierarchical} employ hidden states in a hierarchical framework for long-horizon tasks. However, implicit methods often require retraining in changing environments, limiting adaptability across tasks.

In contrast, explicit storage methods use structured representations like occupancy grids \cite{huang2023fael, cao2021tare, zhang2024efp} and topological maps \cite{chaplot2021seal, xie2022vision, zhang2024hierarchical} to record visited locations. For example, Chaplot et al. \cite{chaplot2021seal} utilize semantic topological maps for object-centric exploration, while Zhang et al. \cite{zhang2024hierarchical} integrate hierarchical reasoning for household object searches. Papatheodorou et al. \cite{papatheodorou2023finding} enhance adaptability in unfamiliar environments by combining semantic exploration with occupancy mapping. The proposed GET also employs explicit storage methods, but unlike these approaches, it uses the probabilistic method that can record multiple locations of the target present in the environment.

Hybrid approaches merge explicit and implicit mechanisms to balance structured clarity with adaptive flexibility. GOAT \cite{chang2023goat} and GOAT-Bench \cite{khanna2024goat} utilize semantic maps and neural network hidden states for multi-modal queries and dynamic adaptation. NaviLLM \cite{zheng2024towards} integrates semantic maps with neural memory, improving multi-modal navigation. Honerkamp et al. \cite{honerkamp2024language} employ open-vocabulary scene graphs for interactive object search, and Dorbala et al. \cite{dorbala2024can} guide zero-shot navigation through explicit semantic graphs and LLM reasoning. While hybrid methods are effective, their tight coupling can make them difficult to extend.

\section{Goal-directed Search Strategy}
\label{sec:search_strategy}
This section introduces the goal-directed search strategy employed in the GET framework, which combines reasoning-based search using LLMs with its core module, DoUT, and experience-based repeated search that utilizes historical data to enhance the efficiency and adaptability of robotic target search tasks. It also discusses key components, such as environmental representation, task probability mapping, and search trajectory refinement, which are primarily utilized in both search methods to ensure smooth and efficient robot movement, enabling real-time adaptation to dynamic environmental changes.

\subsection{Semantic Octomap Environment Representation}
\label{subsec:semantic_octomap}

The Semantic Octomap is used to record the latest environment, enabling quick localization of target positions during repeated search tasks. To construct an accurate semantic octomap, the environmental panorama is first synchronized with the LiDAR point cloud in real-time. Then, it is automatically semantically segmented using Grounded-SAM \cite{ren2024grounded} and RAM \cite{zhang2024recognize}, which generates semantic masks and assigns labels to objects. After segmentation, the point cloud is transformed and projected onto an equidistant cylindrical projection aligned with the panorama, achieving precise pixel-level alignment to create a semantic point cloud. Finally, this semantic point cloud is incrementally mapped into the semantic octomap. To address sensor noise and potential inaccuracies in the automatic labeling process, the construction follows the method proposed by \cite{asgharivaskasi2023semantic}, which employs Bayesian probability to update both occupancy and semantic labels of the octomap cells, allowing for incremental construction and ensuring more accurate environmental representations over time.

\subsection{Task Probability Map Experience Modeling}
Given that objects are typically located in specific positions, a probabilistic approach is advantageous for representing the likelihood of finding objects based on historical search tasks. In the proposed framework, Gaussian Mixture Models (GMMs) \cite{reynolds2009gaussian} are employed to analyze past experiences and capture the probabilities of target locations, with each Gaussian component representing the likelihood of finding the target at a specific location. The Task Probability Map consists of multiple GMMs, each corresponding to the historical positions of a specific search target. This model is particularly valued for its ability to facilitate incremental updates, ensuring data remains up-to-date without extensive reprocessing, as well as for its compact structure, which significantly reduces storage requirements. The detailed introduction to the GMM and its update process are presented in Sec. \ref{sec:task_map}.

\subsection{Diagram of Unified Thought}
DoUT is a general-purpose reasoning enhancement module designed for real-time robotic applications, inspired by DoT \cite{zhang2024diagram}, as illustrated in the upper right corner of Fig. \ref{fig:framework} (more details see Sec. \ref{sec:comparition_dout_dot}). It introduces multiple specialized roles, enabling more refined and nuanced logical reasoning, which makes it effective for handling complex decision-making tasks. Specifically, the LLM acts as a Proposer, combining multimodal environmental inputs and task descriptions to infer potential locations of search targets and generate a series of ranked candidate propositions as selectable decision sequences. An Evaluator operates separately from the LLM, accessing external resources such as octree maps, and utilizes criteria to perform a secondary ranking of the candidate propositions provided by the Proposer, compensating for the Proposer's weak 3D spatial awareness. By conducting parallel assessments of candidate propositions generated by the LLM, the Evaluator consolidates these evaluations into a ranked list of propositions with corresponding scores. These ranked propositions not only guide the robot’s decision-making process but also provide structured feedback to the LLM, fostering continuous improvement in reasoning and learning.

The Evaluator applies two types of evaluation criteria: Mandatory Criteria and Advisory Criteria. Mandatory Criteria include standards such as proposition format and anomaly detection. Candidate propositions that violate any Mandatory Criteria are immediately flagged, and feedback is provided to the Proposer to enforce necessary modifications. Only propositions meeting these criteria proceed to further evaluation. Advisory Criteria focus on ranking candidate propositions and typically include factors such as security, efficiency, and task-specific objectives. These criteria implicitly guide the LLM’s learning process, enhancing its reasoning accuracy and adaptability over time. 

\begin{figure}[tbp]
    \centering
    \includegraphics[width=1.0 \linewidth]{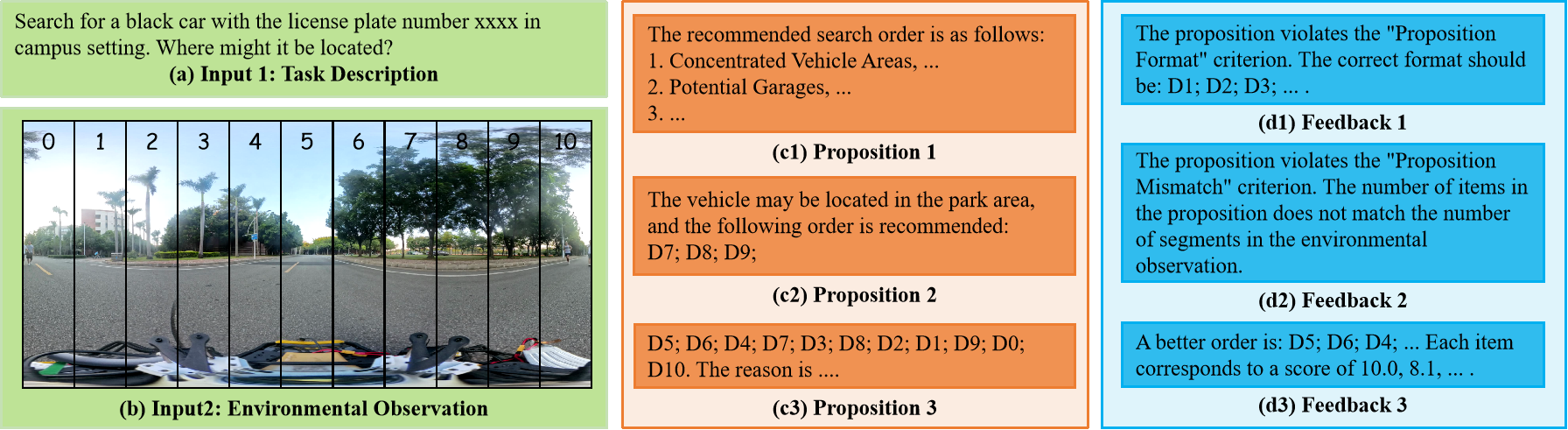}
    \caption{An example of a DoUT-based search task. (a) and (b) present the inputs to DoUT: the task description and environmental panorama. (c1–c3) show candidate propositions generated by the Proposer, while (d1–d3) provide corresponding feedback from the Evaluator. Feedback in (d1–d2) indicates violations of Mandatory Criteria and is returned directly to the Proposer for corrections, whereas (d3) demonstrates valid feedback evaluated by Advisory Criteria.}
    \label{fig:first_search}
\end{figure}

\subsection{Reasoning Search}
\label{subsec:reasoning_search}
At each timestep $t$, an environmental panorama $O_t$ is captured by the panoramic camera at the robot's position $\mathbf{p}_t$. This observation is divided into multiple segments, denoted as $\mathcal{D}_t = \{ D_0, D_2, \cdots, D_{N-1} \}$, as illustrated in Fig.~\ref{fig:first_search}(b). Each segment represents a potential movement direction, facilitating reasoning by the LLM and improving the interpretability of its decisions. To avoid oscillatory behavior during decision-making, it is recommended that $N = 2k + 1$, where $k \in \mathbb{Z}^{+}$. This configuration ensures that the robot's current orientation lies within the middle segment, $D_{(N-1)/2}$. Conversely, when $N$ is even, the robot's orientation falls at the junction of two middle segments, $D_{N/2-1}$ and $D_{N/2}$, potentially causing oscillation between these segments and disrupting smooth movement. This intuitive observation format enables the LLM to reason directly from the current state without the need for complex coordinate transformations, thereby enhancing reasoning accuracy and simplifying the decision-making process.

Subsequently, the task description (Fig.~\ref{fig:first_search}(a)) and the panorama (Fig.~\ref{fig:first_search}(b)) are input into DoUT. The Proposer then utilizes its reasoning capabilities to infer the direction most likely to contain the searched object, outputting candidate proposition list $\mathcal{D}$. However, at the start of the task, the Proposer in DoUT has not undergone task-specific training, and its performance in the search task may still require improvement. Additionally, the environmental panorama is captured in equidistant cylindrical projection, which introduces projection distortion, making it challenging for the Proposer to intuitively interpret distances to surrounding obstacles. Furthermore, DoUT is unaware of the specific output format required to meet the robot’s decision-making needs. Therefore, the Mandatory Criteria and Advisory Criteria of the Evaluator are specifically designed to enhance the reasoning process, resulting in ranked proposition list $\mathcal{D}^*$. Fig.~\ref{fig:first_search} illustrates the feedback and optimization from these criteria on various Proposer propositions, with more details available in Sec. \ref{sec:criteria_details}.

Once the ranked propositions $\mathcal{D}^*$ are determined, they provide a prioritized set of potential directions for the robot’s movement. However, these directions serve merely as broad indications rather than precise target points, therefore, to translate these propositions into an actionable search path, a local target must be selected within the designated direction $D^* \in \mathcal{D}^*$. The selection process follows the rules of Top-ranked Direction, Reachability, and Continuous Movement to ensure that a suitable local target point is identified, with more details available in Sec. \ref{sec:search_path_rules}.

After the local target is determined, the robot evaluates the feasibility of a direct path. If a direct path is obstructed, the robot dynamically searches for an alternative path within the semantic octomap using the A* algorithm. Finally, the path undergoes further refinement to generate a safe, smooth, and continuous trajectory, enabling the robot to search efficiently.

\paragraph{Limitations} Notably, since reasoning search cannot ensure complete coverage of all unexplored areas, once the coverage of the unexplored regions exceeds a certain threshold, the robot queries all unexplored areas from the sparse roadmap. Each of these areas is formulated as a Traveling Salesman Problem (TSP) similar to Eq. \eqref{eq:tsp2}, allowing the robot to determine a route for sequentially visiting the unknown regions. Subsequently, within each area, the robot employs the same strategy as FAEL \cite{huang2023fael} to ensure that all potentially target-containing regions are covered.
\label{sec:limitations}

\subsection{Experienced Search}
When repeated search tasks are performed in the same environment, both the semantic octomap and the multi-layer task probability maps are utilized to enhance search efficiency. During the search process, the GET framework queries these maps to identify potential target locations, formulating a TSP to plan a search tour that efficiently passes through multiple potential target regions.

\subsubsection{Target Locations from Task Probability Maps}
If the multi-layer task probability maps contain the target object, the components of the corresponding GMM are used to generate a set of potential target locations, denoted as $\mathcal{T} = \{T_0, T_1, T_2, \cdots, T_n \}$, where $T_0$ represents the robot's current position $\mathbf{p}_t$, and $T_i$ ($i > 0$) represents the spatial coordinates of the $i$-th Gaussian component. The TSP is then formulated as:
\begin{equation}
\min_{\mathcal{T}} \sum_{(T_i, T_j) \in \mathcal{T}} d(T_i, T_j) \cdot (1 - \beta \pi_j),
\label{eq:tsp1}
\end{equation}
where $d(T_i, T_j)$ represents the distance between two target locations, evaluated using A* in the semantic octomap. $\pi_j$ is the weight of the $j$-th Gaussian component, and $\beta$ is a scaling factor.

\subsubsection{Target Locations from Semantic Octomap}
If the multi-layer task probability maps do not contain the target object, potential target locations $\mathcal{T} = \{ T_0, T_1, T_2, \cdots, T_n \}$ are defined, where $T_0$ represents the robot's current position $\mathbf{p}_t$, and $T_i$ ($i > 0$) are directly identified from the semantic octomap by querying regions labeled with the target category. In this case, the TSP is formulated as:
\begin{equation}
\min_{\mathcal{T}} \sum_{(T_i, T_j) \in \mathcal{T}} d(T_i, T_j).
\label{eq:tsp2}
\end{equation}

Once the TSP is solved, an optimal search tour $\mathcal{T}^{*}$ is obtained. The robot can then search through the semantic octomap to find a path that sequentially passes through the locations in $\mathcal{T}^{*}$, guiding it through the potential target regions for efficient search. If the target is not located after completing the tour, the robot transitions to the Reasoning Search until the target is found.

% \begin{figure}[tbp]
%     \centering
%     \includegraphics[width=0.6 \linewidth]{figures/robot_scene.png}
%     \caption{Experimental scenes and the two-wheeled self-balancing robot used in the experiments.}
%     \label{fig:setup}
% \end{figure}

\begin{figure}[t]
    \centering
    \subfigure [] {
        \includegraphics[width=0.3 \linewidth]{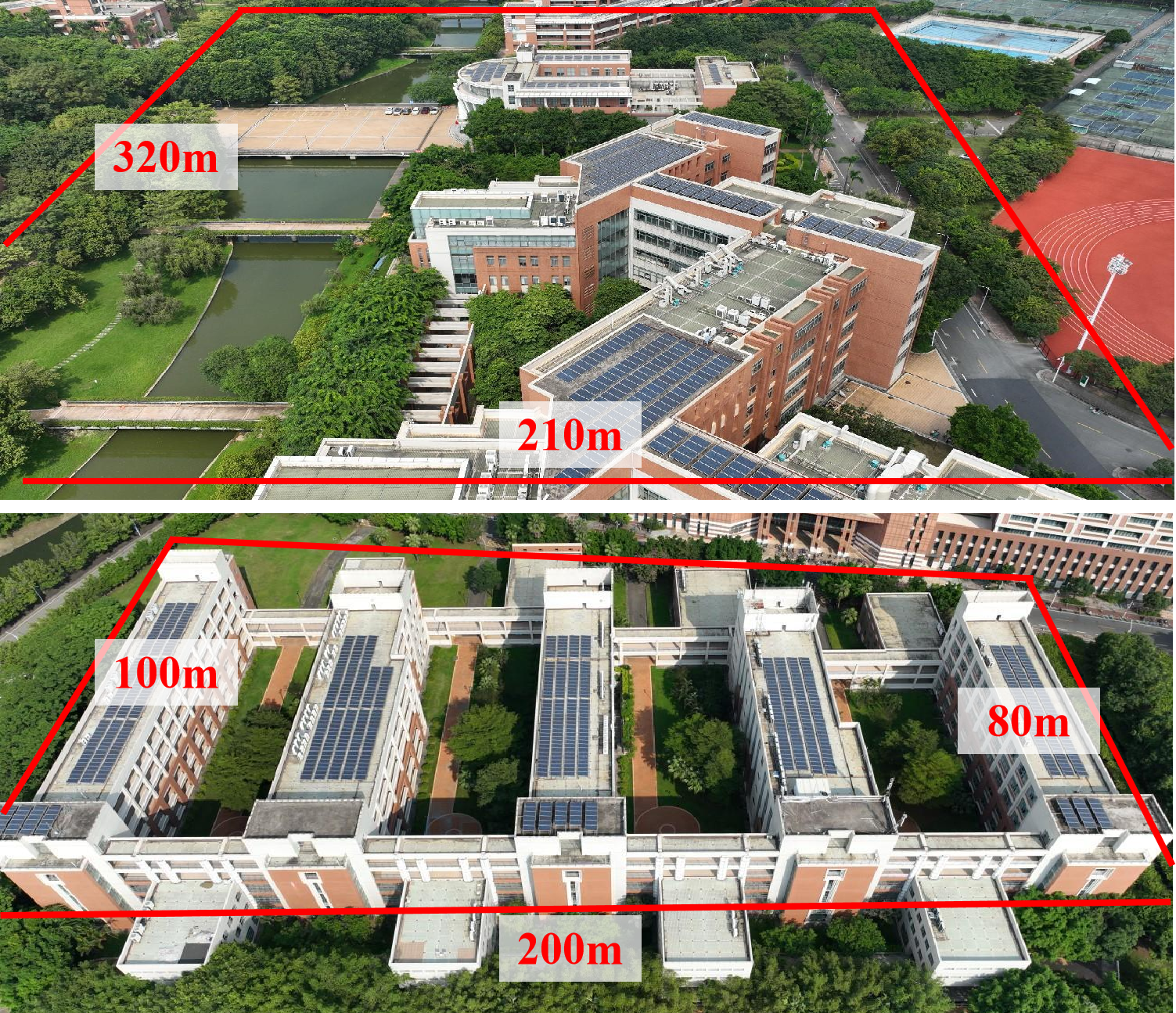}
    }
    \subfigure [] {
        \includegraphics[width=0.2 \linewidth]{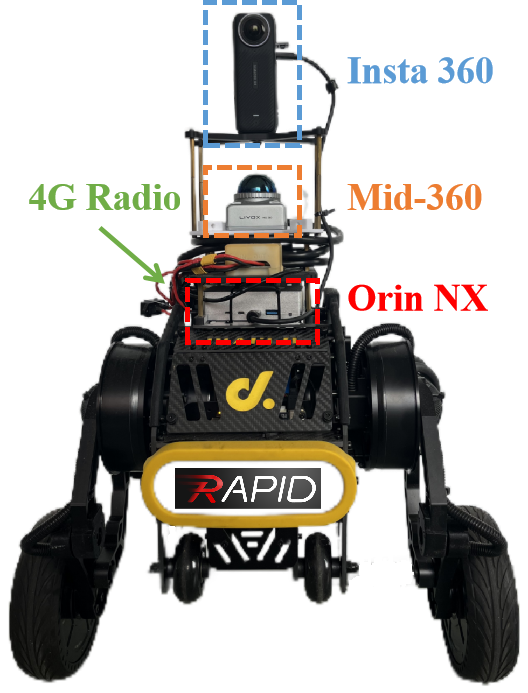}
    }
    \caption{Experimental setups for the robot and real-world scenes.}
    \label{fig:setup} 
\end{figure}

\subsection{Search Trajectory Refinement} 
Once the initial search path is determined, it is further optimized to ensure dynamic feasibility and account for environmental constraints. While the A* algorithm provides a collision-free path, it often neglects the robot's dynamic capabilities, resulting in sharp turns or highly angular trajectories. Such paths may require the robot to decelerate significantly or execute large-angle rotations, reducing overall efficiency. Moreover, dynamic obstacles in the environment, such as pedestrians or vehicles, necessitate continuous adjustments to maintain safe and efficient navigation.

To address these challenges, the Trajectory Refinement module employs a fixed-size Euclidean Signed Distance Field (ESDF) that dynamically updates as the robot moves. The ESDF is periodically cleared and rebuilt to maintain accuracy, particularly after the robot has traveled a certain distance, ensuring that newly detected dynamic obstacles are incorporated. The search path is continuously refined within the ESDF to ensure real-time feasibility and adaptability. Additionally, the path is further optimized using the method described in \cite{zhou2019robust}, which enhances both safety and smoothness while seamlessly adapting to updated local targets and decisions.

% **********************************************

\section{Experiments}
\label{sec:experiments}
% This section outlines the experimental setup and presents results that evaluate search efficiency through various metrics, confirming the proposed framework's effectiveness.

\subsection{Implementation Details}
Experiments are conducted in two large-scale environments, as shown in Fig.~\ref{fig:setup} (a). Scene 1 is an outdoor campus area measuring $320 \text{m} \times 210 \text{m}$, featuring moving vehicles and pedestrians, with the goal of searching for a target vehicle. Scene 2 consists of four interconnected semi-open buildings, spanning $200 \text{m} \times 100 \text{m}$, where the objective is to locate a specific classroom. A two-wheeled self-balancing robot, depicted in Fig.~\ref{fig:setup} (b), is used for the experiments. It operates at a maximum speed of $1.0 \text{ m/s}$ and is equipped with an NVIDIA Jetson Orin NX for real-time processing. Environmental data is captured using an Insta360 X4 camera, providing 360-degree images at a resolution of 2448×1224 and 30 fps, and a Mid-360 LiDAR, delivering point clouds at 10 Hz. A 4G radio ensures network connectivity with upload speeds of 50 Mbps and download speeds of 150 Mbps. High-precision localization is achieved using FAST-LIO2 \cite{fast_lio}. The semantic octomap has a cell size of $0.1 \text{m}$. GPT-4o-mini \cite{achiam2023gpt} serves as the LLM, supporting simultaneous image and text input. Path optimization is conducted with a maximum acceleration of $1 \text{ m/s}^2$ and an angular velocity limit of $1 \text{ rad/s}$, while maintaining a $1 \text{ m}$ safety distance from obstacles. The environmental panorama is divided into 11 segments for decision-making, with evaluation weights set as $\lambda_1 = 2.5$, $\lambda_2 = 10.0$, $\lambda_3 = 3.0$, and $\lambda_4 = 1.5$, emphasizing safety, minimizing redundancy, and ensuring smooth movement.

\subsection{Search Results and Analysis}
Fig.~\ref{fig:search_trajectories} compares the robot's search trajectories across two distinct scenes with benchmark methods, including Sem \cite{papatheodorou2023finding}, FAEL \cite{huang2023fael}, EFP \cite{zhang2024efp}, and TARE \cite{cao2021tare}. All benchmark implementations follow identical hardware setups and parameter settings, adhering to the recommendations provided in their respective open-source projects. The colored point cloud in Fig.~\ref{fig:search_trajectories} represents areas traversed by the proposed method and recorded within the semantic octomap, while unexplored regions are depicted in gray. As shown in Fig.~\ref{fig:search_trajectories}, GET demonstrates shorter trajectories in both inference search and repeated search tasks compared to the benchmarks, due to DoUT's ability to infer potential target locations, allowing the robot to explore these areas more efficiently and avoid excessive movement, thereby significantly improving exploration efficiency.

\subsubsection{Reasoning Search}
\begin{figure*}[t]
    \centering
    \subfigure[] {
        \includegraphics[scale=0.245]{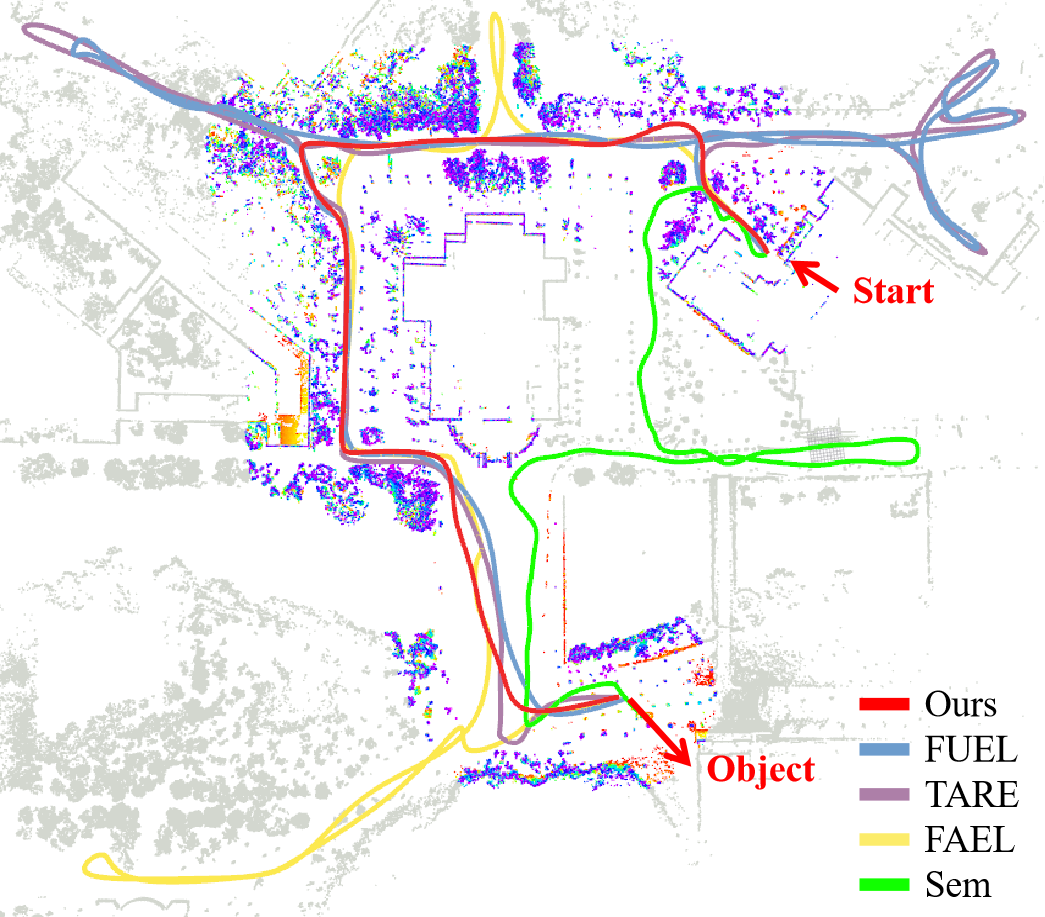}
    }
    \subfigure[] {
        \includegraphics[scale=0.235]{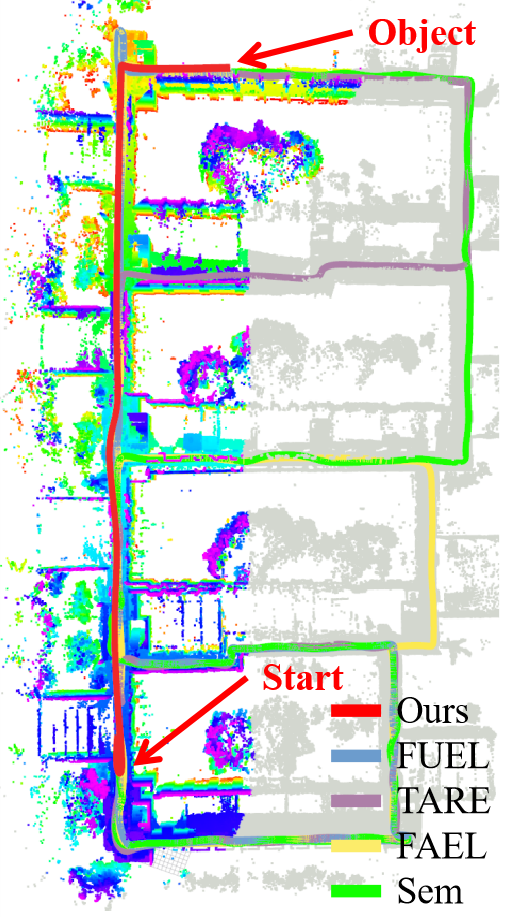}
    }
    \subfigure[] {
        \includegraphics[scale=0.245]{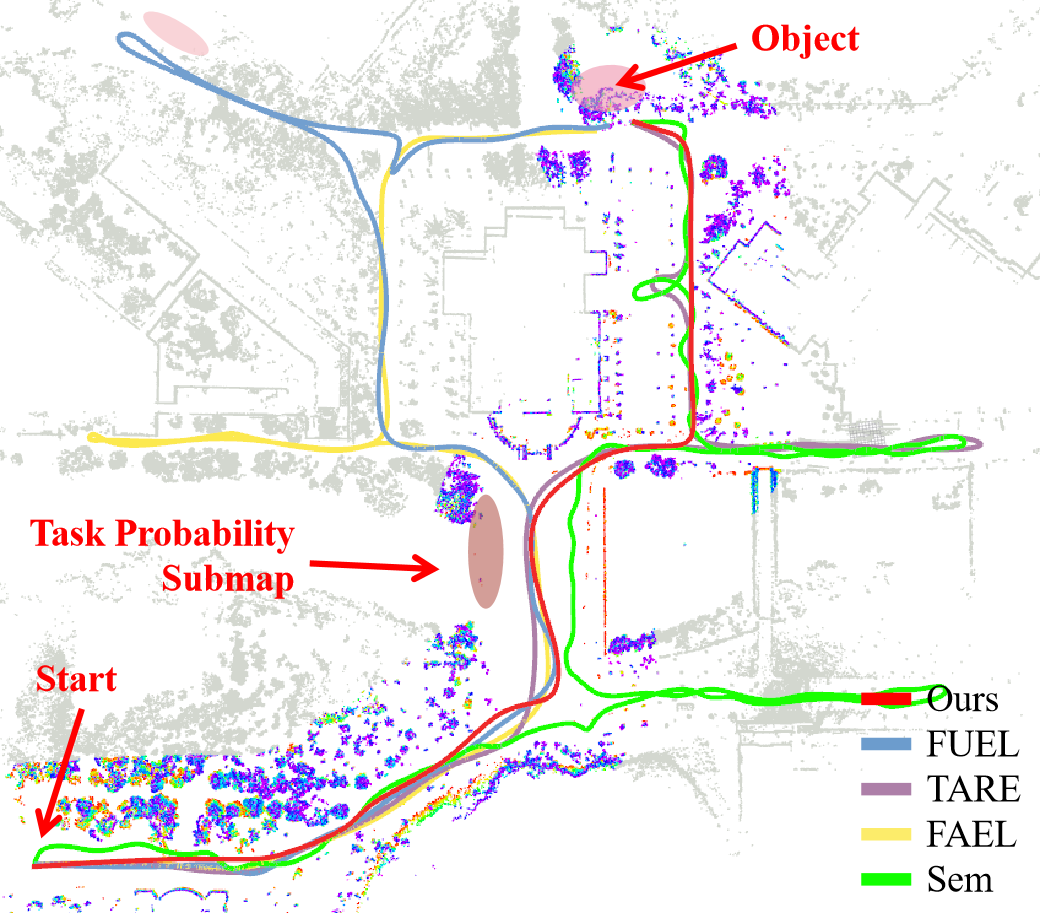}
    }
    \subfigure[] {
        \includegraphics[scale=0.235]{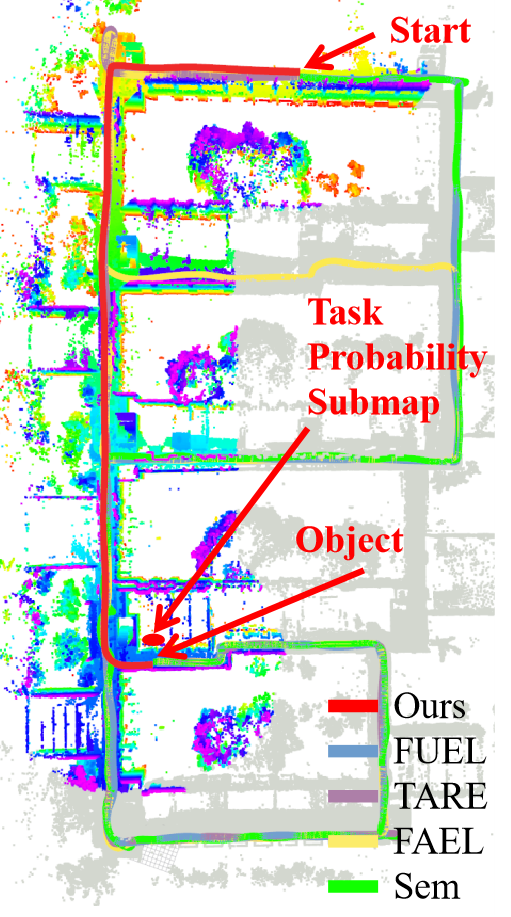}
    }
    \caption{Comparison of the proposed algorithm's search trajectories against benchmarks. (a) and (b) illustrate searches without prior experience, while (c) and (d) show experienced searches utilizing historical search experience, with variations in both the starting position and the search object.}
    \label{fig:search_trajectories} 
\end{figure*}

In the first-time search task, due to a lack of environmental information, GET employs reasoning search. As shown in Figs.~\ref{fig:search_trajectories}(a) and (b), the proposed method demonstrates superior performance with the shortest search trajectories compared to the benchmarks, owing to the LLM's strong environmental comprehension and reasoning capabilities, which prioritize areas more likely to contain the target. Furthermore, the evaluator within DoUT accesses the sparse roadmap and, through non-mandatory criteria, not only prevents collisions caused by the LLM's limited 3D spatial awareness but also avoids decision conflicts by steering the robot's decisions away from already explored areas. Sec. \ref{sec:dout_evaluation} provides more evaluations of DoUT and details on the reasoning search. As a result, the robot can continuously explore ahead, avoiding unnecessary wandering within local regions. Table~\ref{tab:reasoning_search} provides detailed comparisons of the search trajectories and efficiency metrics.
 
Although all methods achieve a 100\% success rate in locating the target across multiple experiments, significant differences are observed in terms of search paths and efficiency. In Scene~1, GET  achieves an average path length of 390.41~m, reducing the path length by 45.8\%, 54.2\%, 57.2\%, and 25.2\% compared to FAEL, TARE, EFP, and Sem, respectively. Similarly, the search time is reduced to 412.40~s, with improvements of 48.1\%, 58.7\%, 58.1\%, and 29.1\% over the same benchmarks. In Scene~2, GET  achieves even greater efficiency, with an average path length of 191.97~m and a search time of 199.97~s, significantly outperforming the benchmarks. These results demonstrate the ability of GET  to leverage the LLM’s reasoning and environmental understanding to minimize both path length and search time, even without prior information. 

\begin{table}[t]
\centering
\small
\setlength{\tabcolsep}{2.1pt}
\caption{Comparison of first-time search in two scenes}
\begin{tabular}{ccccccc|ccccc}
\hline \hline
\multicolumn{1}{l}{}                                                                 & \multicolumn{1}{l}{} & \multicolumn{5}{c|}{\textbf{Scene 1}} & \multicolumn{5}{c}{\textbf{Scene 2}} \\ \hline
\textbf{}                                                                            & \textbf{}            & FAEL  & TARE  & EFP   & Sem   & GET  & FAEL   & TARE  & EFP  & Sem & GET  \\ \hline
\multirow{2}{*}{\textbf{\begin{tabular}[c]{@{}c@{}}Path \\ Length (m)\end{tabular}}} & Avg                  & 720.32 & 852.53 & 912.50 & 521.81 & \textbf{390.41} & 380.60 & 481.09 & 496.74 & 441.10 & \textbf{191.97} \\
                                                                                     & Std                  & 48.43  & 65.82  & 49.24  & 68.13 & \textbf{15.39}  & 23.15  & 26.07  & 34.29  & 29.98  & \textbf{8.08}  \\ \hline
\multirow{2}{*}{\textbf{\begin{tabular}[c]{@{}c@{}}Time (s)\end{tabular}}}         & Avg                  & 794.47 & 999.06 & 984.71 & 581.34 & \textbf{412.40} & 422.89 & 559.41 & 521.06 & 479.77 & \textbf{199.97} \\
                                                                                     & Std                  & 40.89  & 52.47  & 41.63  & 46.78 & \textbf{10.58}  & 18.34  & 21.92  & 27.65  & 24.33  & \textbf{6.78}  \\ \hline
\end{tabular}
\label{tab:reasoning_search}
\end{table}

\subsubsection{Experienced Search}
In the repeated search tasks, the robot leverages multi-layer task probability maps and the semantic octomap to further enhance search efficiency. As shown in Figs.~\ref{fig:search_trajectories}(c) and (d), the Gaussian components of the task probability map are visualized in varying shades of red, with darker shades indicating higher probabilities of locating the target. 

By systematically prioritizing regions with higher probabilities based on prior search data or semantic octomap information, the robot effectively navigates complex environments and dynamically adjusts its search strategy. As a result, the robot achieves shorter and more optimized trajectories compared to the benchmarks. Even when starting positions or target locations vary across tasks, the proposed method demonstrates robust adaptability, consistently maintaining high search efficiency.

\begin{table}[]
\centering
\small
\caption{Comparison of repeated search in two scenes}
\begin{tabular}{cccccccccc}
\hline
                                  &                                                      & \multicolumn{4}{c}{\textbf{Experienced Search - Same Task}}           & \multicolumn{4}{c}{\textbf{Experienced Search - Changed Task}}        \\ \cline{3-10} 
                                  & \textbf{Method}                                      & \multicolumn{2}{c}{Path Length (m)} & \multicolumn{2}{c}{Time (s)}    & \multicolumn{2}{c}{Path Length (m)} & \multicolumn{2}{c}{Time (s)}    \\ \cline{3-10} 
                                  &                                                      & Avg               & Std             & Avg             & Std           & Avg               & Std             & Avg             & Std           \\ \hline
\multirow{5}{*}{\textbf{Scene 1}} & FAEL \cite{huang2023fael}           & 720.32            & 48.43           & 794.47          & 40.89         & 593.80            & 44.23           & 650.15          & 67.45         \\
                                  & TARE \cite{cao2021tare}             & 852.53            & 65.82           & 999.06          & 52.47         & 604.61            & 57.56           & 714.11          & 52.67         \\
                                  & EFP \cite{zhang2024efp}            & 912.50            & 49.24           & 984.71          & 41.63         & 561.59            & 43.89           & 610.42          & 49.81         \\
                                  & Sem \cite{papatheodorou2023finding} & 521.81            & 68.13           & 581.34          & 46.78         & 494.92            & 55.61           & 552.92          & 58.28         \\
                                  & GET                                                 & \textbf{353.67}   & \textbf{5.36}   & \textbf{363.36} & \textbf{4.72} & \textbf{378.58}   & \textbf{11.47}  & \textbf{383.70} & \textbf{8.34} \\ \hline
\multirow{5}{*}{\textbf{Scene 2}} & FAEL \cite{huang2023fael}           & 380.60            & 23.15           & 422.89          & 18.34         & 451.40            & 28.65           & 497.86          & 12.56         \\
                                  & TARE \cite{cao2021tare}             & 481.09            & 26.07           & 559.41          & 21.92         & 404.49            & 36.34           & 470.34          & 11.29         \\
                                  & EFP \cite{zhang2024efp}            & 496.74            & 34.29           & 521.06          & 27.65         & 613.20            & 35.78           & 647.75          & 24.90         \\
                                  & Sem \cite{papatheodorou2023finding} & 441.10            & 29.98           & 479.77          & 24.33         & 710.72            & 48.32           & 757.44          & 51.05         \\
                                  & GET                                                 & \textbf{179.88}   & \textbf{3.91}   & \textbf{185.19} & \textbf{3.15} & \textbf{186.19}   & \textbf{6.92}   & \textbf{193.95} & \textbf{6.78} \\ \hline
\end{tabular}
\label{tab:repeated_search}
\end{table}

Table~\ref{tab:repeated_search} presents the execution of repeated search tasks in different scenarios. In this context, \textit{Experienced Search - Same Task} refers to situations where both the starting position and target location remain unchanged, while \textit{Experienced Search - Changed Task} indicates that both the starting position and target location vary. Benefiting from accumulated historical data, the proposed method improves its performance over time. For instance, in Scene~1, the average path length decreases from 390.41 m in the first-time search to 353.67 m in experienced searches with the same task, while the standard deviation drops from 15.39 to 5.36. Similarly, the search time decreases from 412.40 s to 363.36 s, accompanied by a reduction in standard deviation from 10.58 to 4.72. In Scene~2, both path length and search time further reduce to 179.88 m and 185.19 s, respectively. Compared to the benchmarks, GET consistently demonstrates superior performance and adaptability across scenarios.

When changing starting positions and targets (\textit{Experienced Search - Changed Task}), the proposed method achieves significant efficiency gains. For example, in Scene~1, the path length is reduced to 378.58~m, while the search time decreases to 383.70~s. In Scene~2, the method maintains a similarly strong performance, with a path length of 186.19~m and a search time of 193.95~s. These results underscore the robustness and adaptability of the proposed method in dynamically adjusting search strategies and maintaining high efficiency even under changing conditions. 

\begin{table}[]
\small
\caption{Average time consumption of each module}
\centering
\begin{tabular}{ccccc}
\hline \hline
\begin{tabular}[c]{@{}c@{}}Panoramic\\ Segmentation (ms)\end{tabular} & \begin{tabular}[c]{@{}c@{}}Semantic\\ Match (ms)\end{tabular} & \begin{tabular}[c]{@{}c@{}}Network\\ Delay (ms) \end{tabular} & \begin{tabular}[c]{@{}c@{}}GPT-4\\ Response (ms)\end{tabular} & \begin{tabular}[c]{@{}c@{}}Trajectory\\ Optimization (ms)\end{tabular} \\ \hline
1850.47 & 6.87 & 43.15 & 527.82 & 8.64 \\ \hline
\end{tabular}
\label{tab:time_consumption}
\end{table}

Table~\ref{tab:time_consumption} presents the average time consumption of each module within the proposed framework. While some modules, such as panoramic segmentation and GPT-4 decision-making, exhibit relatively longer processing times, the framework ensures real-time operation through several carefully designed strategies. During short-distance movements, the robot's visibility of the point cloud remains largely unaffected by occlusions. As a result, the semantic matching of point clouds with panoramic images, performed at approximately 2~s intervals, does not significantly impact overall performance. The GET framework effectively leverages the LLM's robust environmental understanding and accurate inference capabilities, combined with trajectory optimization and other modules, allowing it to maintain real-time performance even when certain modules require longer processing times.

\section{Conclusion}
This paper introduces GET, a novel framework that leverages the reasoning and understanding capabilities of LLMs for autonomous target searches in large-scale, unknown environments. By incorporating the DoUT module, GET significantly enhances reasoning accuracy and decision-making, enabling autonomous task-specific learning. The framework integrates multi-layer task probability maps, progressively optimizing search strategies based on historical task experiences. Experimental results consistently demonstrate that GET outperforms benchmark methods, achieving substantial reductions in path length and search time, even under varying starting positions and target locations. Extensive testing across multiple LLMs underscores the exceptional adaptability and robust performance of the DoUT module, positioning GET as an efficient and scalable solution for real-time robotic decision-making.

% \bibliographystyle{Ref}  %plainnat,abbrvnat,unsrtnat
% \small
% \bibliography{reference}
% \normalsize

\bibliographystyle{unsrt}
% \setcitestyle{square,numbers,comma}
\bibliography{reference}

% \section*{References}

%%%%%%%%%%%%%%%%%%%%%%%%%%%%%%%%%%%%%%%%%%%%%%%%%%%%%%%%%%%%

\appendix

\section{Technical Appendices and Supplementary Material}

\subsection{Task Probability Map}
\label{sec:task_map}
The Task Probability Map utilizes the Gaussian Mixture Model (GMM) at each layer to represent the probability distribution of target locations, with each GMM corresponding to a specific search target in the environment. This probabilistic approach allows for an effective representation of uncertainty and variability in target positions, enabling the system to adapt to new information as it becomes available. By incorporating historical data and dynamically updating the model, the GMM facilitates informed decision-making during search tasks. The following sections detail the formulation of the GMM, the process of creating and updating Gaussian components, and the mechanisms that ensure the model remains responsive to changes in the environment.
\subsubsection{Gaussian Mixture Model}
The GMM \cite{reynolds2009gaussian} is a probabilistic model that represents a mixture of multiple Gaussian distributions. Each component of the GMM corresponds to a Gaussian distribution, representing the probability of encountering the target at specific locations. The probability density function for each component is defined as:
\begin{equation} 
\mathcal{N}(\mathbf{x}|\mu_i, \Sigma_i) = \frac{1}{\sqrt{(2\pi)^k |\Sigma_i|}} e^{-\frac{1}{2} (\mathbf{x} - \mu_i)^T \Sigma_i^{-1} (\mathbf{x} - \mu_i)},
\end{equation}
where $\mathbf{x} \in \mathbb{R}^k$ represents the spatial coordinates, $\mu_i$ is the mean of the $i$-th Gaussian component, indicating the observed object location, and $\Sigma_i$ is the covariance matrix, quantifying the uncertainty in the object's position.

For a GMM consisting of $m$ components, the probability of finding the object at a specific location $\mathbf{x}$ is the weighted sum of the probabilities provided by each Gaussian component, , with $\pi_i$ denoting the weight of the $i$-th component:
\begin{equation}
    p(\mathbf{x} | \{\pi_i, \mu_i, \Sigma_i \}_{i=1}^m) = \sum_{i=1}^{m} \pi_i \mathcal{N}(\mathbf{x}|\mu_i, \Sigma_i).
\label{eq:probability}
\end{equation}

\subsubsection{Creating and Updating Gaussian Components}
When the search task is successfully completed and the target object is found at position $\mathbf{p}$, the GMM is updated using the following steps:

\paragraph{1) Adding a New Gaussian Component}
A new Gaussian component $G_{\text{new}} = \{\pi_{\text{new}}, \mu_{\text{new}}, \Sigma_{\text{new}} \}$ is added to the model. Its parameters are defined as:
\begin{equation}
\begin{aligned}
    \mu_{\text{new}} &= \mathbf{p}, \\
    \Sigma_{\text{new}} &= \sigma^2 \mathbf{I}, \\
    \pi_{\text{new}} &= \frac{1}{m+1},
\end{aligned}
\end{equation}
where $m$ is the number of existing components in the GMM, and $\sigma^2$ is proportional to the object's size, representing the uncertainty in its location.

\paragraph{2) Merging Gaussian Components}
If the Euclidean distance $\|\mu_{\text{new}} - \mu_i\|$ between the new component $G_{\text{new}}$ and an existing component $G_i$ is smaller than the sum of their standard deviations, they are merged. The updated parameters of the Gaussian component $G_i'$ are:
\begin{equation}
\begin{aligned}
\mu_i' & = \frac{\pi_i \cdot \mu_i + \pi_{\text{new}} \cdot \mu_{\text{new}}}{\pi_i + \pi_{\text{new}}}, \\
\Sigma_i' & = \frac{\pi_i}{\pi_i + \pi_{\text{new}}} \left( \Sigma_i + (\mu_i - \mu_i')(\mu_i - \mu_i')^T \right) + \frac{\pi_{\text{new}}}{\pi_i + \pi_{\text{new}}} \left( \Sigma_{\text{new}} + (\mu_{\text{new}} - \mu_i')(\mu_{\text{new}} - \mu_i')^T \right), \\
\quad \pi_i' & = \pi_i + \pi_{\text{new}}.
\end{aligned}
\end{equation}
After merging, the newly created component $G_{\text{new}}$ is discarded. This process combines historical and recent data, providing a more accurate reflection of the object's probability distribution.

\paragraph{3) Gaussian Component Weight Normalization}
The weights of all components are normalized as follows:

\begin{equation}
    \label{eq:update_other_gaussian_weight}
    \pi_j = \frac{\pi_j}{\sum_{k=1}^{m} \pi_k}.
\end{equation}

This process allows the GMM to dynamically adapt to environmental changes. With each new experience, it updates the target object's location while reducing the influence of outdated data. This adaptive approach efficiently represents search experiences using the parameters $\{\pi_i, \mu_i, \Sigma_i\}_{i=1}^m$, thereby lowering the maintenance costs of large-scale environmental maps.

\subsection{Criteria Design Details}
\label{sec:criteria_details}

This section details the design of both Mandatory and Advisory Criteria used to evaluate the quality and relevance of candidate propositions. The Mandatory Criteria establish fundamental requirements that must be satisfied for a proposition to be deemed valid, while the Advisory Criteria introduce additional evaluation layers that refine the decision-making process. By integrating these criteria, the system can effectively rank propositions based on safety, exploration efficiency, and movement consistency, ultimately optimizing the robot's search strategy in dynamic environments.

\subsubsection{Mandatory Criteria}
\label{sec:mandatory_criteria}
\paragraph{1) Proposition Format: }
In its initial attempts, the Proposer may fail to generate propositions in the structured format required by the Evaluator, such as $\mathcal{D}=\{D_0; D_1; D_2; \cdots\}$, as shown in Fig.~\ref{fig:first_search} (c1). In such cases, the Evaluator provides direct feedback specifying the expected format, as shown in Fig.~\ref{fig:first_search} (d1). This process is repeated until the propositions conform to the required structure.

\paragraph{2) Proposition Mismatch: }
If the number of items in the candidate propositions $\mathcal{D}$ generated by the Proposer does not match the number of segments derived from the environmental panorama, the Evaluator flags this inconsistency and provides corrective feedback, as shown in Fig.~\ref{fig:first_search} (c2) and Fig.~\ref{fig:first_search} (d2).

Once all Mandatory Criteria are met, the following Advisory Criteria are applied to further evaluate and rank the propositions $\mathcal{D}$, producing the sorted propositions $\mathcal{D}^*$.

\subsubsection{Advisory Criteria}
\label{sec:advisory_criteria}
\paragraph{1) Security Criteria}
This penalizes propositions that bring the robot closer to obstacles. The real-time point cloud is divided into $N$ regions, each uniquely corresponding to a candidate proposition $D_i \in \mathcal{D}$ from the environmental panorama. For each region, the distance $d_i$ to the nearest obstacle is estimated. The penalty $\mathcal{C}_{s,i}$ for candidate $D_i$ is:

\begin{equation}
    \mathcal{C}_{s,i} = \text{max}(0, d_{\text{safe}} - d_i)^2
\end{equation}
where $d_{\text{safe}}$ is a predefined safety distance. If $d_i \geq d_{\text{safe}}$, no penalty is applied. Otherwise, the penalty grows quadratically as the distance to the nearest obstacle decreases.

\paragraph{2) Repeated Exploration Criteria}
This penalizes propositions that lead the robot toward previously explored areas. To quantify the degree of redundancy, a sparse roadmap \cite{huang2023fael} is constructed for the obstacle-free regions the robot has already traversed. The areas enclosed by the edges of the sparse roadmap provide an estimation of the previously explored regions. The overlap between the area covered by each candidate proposition $D_i$ and the sparse roadmap is then used to compute the penalty $\mathcal{C}_{r,i}$:
\begin{equation}
    \mathcal{C}_{r,i} = \frac{S_{\text{overlap}}}{S_{\text{cover}}}
\end{equation}
where $S_{\text{cover}}$ is the total LiDAR-covered free space, and $S_{\text{overlap}}$ is the overlap area with previously explored regions.

\paragraph{3) Direction Change Penalty}
This penalizes propositions where the direction at timestep $t$ deviates significantly from the robot's movement direction at $t-1$, causing frequent changes in movement direction, which can lead to inefficiencies or excessive energy consumption. The penalty $\mathcal{C}_{d,i}$ for candidate $D_i$ is:
\begin{equation}
    \mathcal{C}_{d,i} = 1 - \cos(\theta_i - \theta_{t-1})
\end{equation}
where $\theta_i$ is the direction of $D_i$, and $\theta_{t-1}$ is the direction at the previous timestep.

The final score $C_i$ assigned by the Evaluator to each candidate proposition $D_i$ is calculated as:
\begin{equation}
    C_i = \lambda_1 \text{Order}(i) + \lambda_2 \mathcal{C}_{s,i} + \lambda_3 \mathcal{C}_{r,i} + \lambda_4 \mathcal{C}_{d,i},
    \label{eq:merge_criteria}
\end{equation}
where $\lambda_1, \lambda_2, \lambda_3, \lambda_4$ are weights that balance the influence of each criterion. $\text{Order}(i)$ denotes the original index of $D_i$ in the list of candidate propositions $\mathcal{D}$ generated by the Proposer, encouraging prioritization based on their initial order.

Finally, each candidate $D_i \in \mathcal{D}$ is evaluated using Eq.~(\ref{eq:merge_criteria}), and the resulting scores $C_i \in \mathcal{C}$ are utilized to sort $\mathcal{D}$ in ascending order, producing the ranked propositions $\mathcal{D}^*$ along with the sorted scores $\mathcal{C}^*$. These ranked propositions $\mathcal{D}^*$ not only serve as the decision basis for the robot's search but also, together with their corresponding scores $\mathcal{C}^*$, are fed back to the LLM to facilitate task-specific learning and improve reasoning accuracy.

\subsection{Search Path Generation Rules}
\label{sec:search_path_rules}
\paragraph{1) Top-ranked direction} The highest-ranked direction $D^*$ is initially chosen as the robot’s intended movement. If $D^*$ fails to meet the criteria below, the next best-ranked direction in $\mathcal{D}^*$ is selected as the new $D^*$.

\paragraph{2) Reachability} The selected direction must lead to a free cell in the semantic octomap, ensuring accessibility to avoid collisions with static or dynamic obstacles. Additionally, the distance from the target cell to the nearest occupied cell must exceed the robot's size, providing sufficient clearance for safe navigation.

\paragraph{3) Continuous Movement}  
The local target should maintain an appropriate distance from the robot's current position, ideally positioned at the center of the designated direction $D^*$. This ensures sufficient travel distance and prevents the robot from halting upon reaching the local target during DoUT reasoning cycles, thereby maintaining continuous movement.

\subsection{Compariton between DoUT and DoT}
\label{sec:comparition_dout_dot}

This section compares the DoT and the DoUT, focusing on their efficiencies and adaptability in robotic applications. While DoT enhances reasoning through specialized roles, it faces challenges such as time consumption and limited flexibility. In contrast, DoUT addresses these limitations with a parallel evaluation mechanism, streamlining decision-making processes. We will explore their structural differences, the challenges of DoT, and the advantages of DoUT in real-time reasoning.

\subsubsection{Diagram of Unified Thought}
\begin{figure}[]
    \centering
    \subfigure [Diagram of Thought (DoT) \cite{zhang2024diagram}: Directed acyclic graph of iterative reasoning.] {
        \includegraphics[scale=0.5]{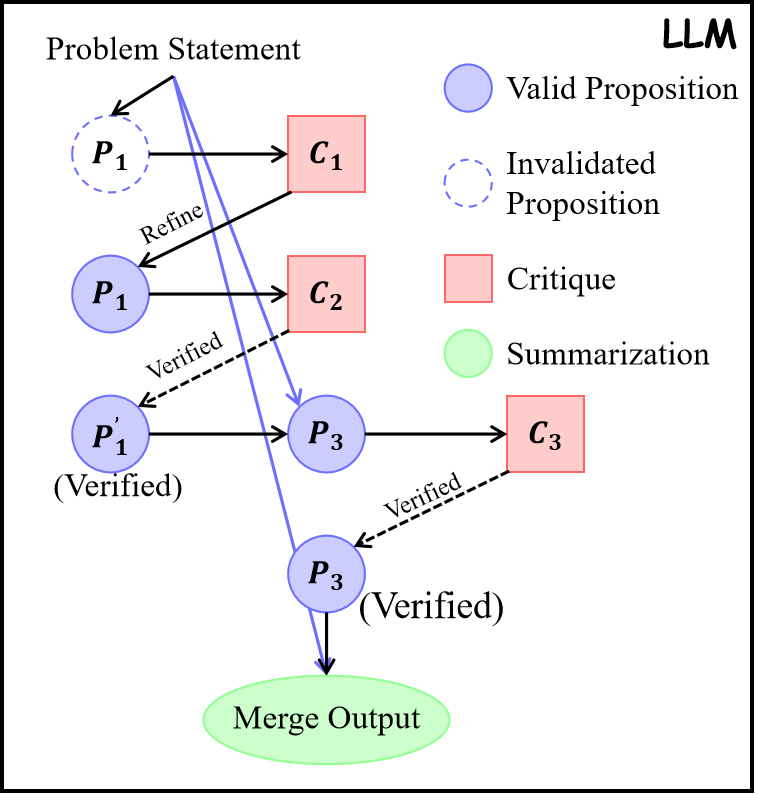}
    }
    \subfigure [Diagram of Unified Thought (DoUT): Directed acyclic graph of reasoning with unified evaluation, feedback, and iteration.] {
        \includegraphics[scale=0.5]{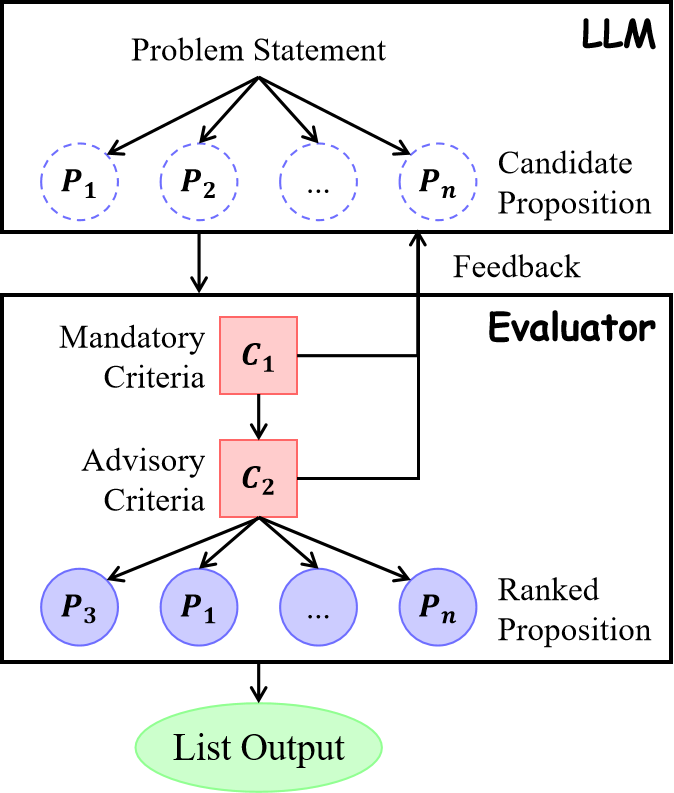}
    }
    \caption{Comparison of DoT and DoUT structures. (a) DoT employs a sequential iterative reasoning process represented by a directed acyclic graph, where the LLM generates propositions, evaluates them using critiques, and refines or verifies outputs based on feedback. (b) DoUT enhances this process with a parallel framework by integrating unified evaluation criteria through an external Evaluator, which ranks candidate propositions from the LLM based on mandatory and advisory criteria, providing structured feedback to improve decision-making and reasoning efficiency.}
    \label{fig:dot_dout} 
    % \vspace{-5mm}
\end{figure}
% \begin{figure}[]
%     \centering
%     \includegraphics[width=0.4 \linewidth]{figures/dot.png}
%     \caption{Diagram of Thought (DoT) \cite{zhang2024diagram}: Directed acyclic graph of iterative reasoning. DoT employs a sequential iterative reasoning process represented by a directed acyclic graph, where the LLM generates propositions, evaluates them using critiques, and refines or verifies outputs based on feedback.}    
%     \label{fig:dot_framework}
% \end{figure}

DoT \cite{zhang2024diagram} is a reasoning framework designed to enhance the capabilities of LLMs by introducing multiple specialized roles: the Proposer, which generates candidate propositions; the Critic, which evaluates and refines these propositions; and the Summarizer, which consolidates verified reasoning steps to produce the final output, as shown in Fig. \ref{fig:dot_dout}(a). This collaborative, multi-role structure mimics human cognitive processes, with distinct components contributing to various stages of reasoning. By leveraging this structure, DoT enables more refined and nuanced logical reasoning, making it effective for handling complex decision-making tasks.

While DoT has proven successful in improving reasoning efficiency with interaction between model components, directly applying it to robotic applications presents several challenges:

\textbf{1) Time-Intensive Reasoning:}
The iterative nature of DoT requires multiple reasoning cycles for each task, introducing delays in decision-making and slowing down the robot’s response time, which limits its real-time adaptability.

\textbf{2) Compromised Integrity:} 
Changing search targets or adapting to new objectives often requires reconfiguring the Critic's evaluation criteria. Such frequent modifications to the internal structure of the LLM risk disrupting its overall integrity, potentially leading to unpredictable and undesirable outcomes, which can significantly hinder its flexibility and reliable deployment across diverse tasks.

\textbf{3) Application-Specific Coupling:}
Many critics rely heavily on external resources, such as maps, requiring frequent access to external memory. This dependence increases system complexity and reduces reasoning efficiency, as the model must continuously retrieve and process external data.

\textbf{4) Challenges in Adaptive Response:}
Robotic applications inherently involve uncertainty \cite{thrun2002probabilistic}, but DoT’s single-reasoning approach lacks the adaptability required for time-sensitive and dynamic environments. This limitation restricts the robot's ability to respond effectively to rapidly changing conditions.

\begin{figure}[tbp]
    \centering
    \includegraphics[scale=0.28]{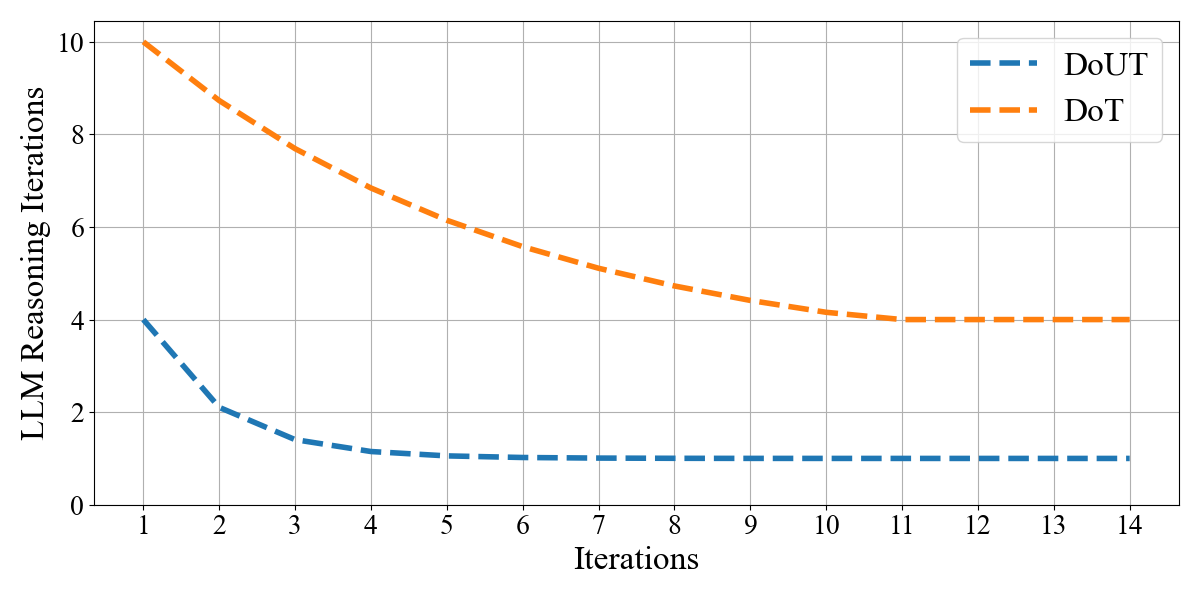}
    \caption{Comparison of LLM reasoning iterations required for DoT and DoUT. DoUT shows a notable decrease in reasoning iterations compared to DoT.}    
    \label{fig:compare_dot_dout}
\end{figure}

To address these challenges, DoUT is designed for real-time robotic applications, inspired by DoT. In DoUT, an Evaluator operates separately from the LLM and replaces the internal Critic and Summarizer roles of DoT, as illustrated in Fig. \ref{fig:dot_dout}(b). By conducting parallel assessments of candidate propositions generated by the LLM, the Evaluator consolidates the evaluations into a ranked list of propositions with corresponding scores. These ranked propositions not only guide the robot’s decision-making process but also provide structured feedback to the LLM, fostering continuous improvement in reasoning and learning. The advantages of DoUT are summarized as follows:

\textbf{1) Integrity:}
By centralizing the evaluation process within an external Evaluator, DoUT eliminates the need for internal modifications of the LLM’s evaluation criteria, thereby preserving its integrity.

\textbf{2) Decoupling: }
The Evaluator centralizes and standardizes memory access, avoiding direct access by the LLM. This design achieves decoupling, reduces system complexity, and mitigates risks associated with unauthorized or inconsistent memory access.

\textbf{3) Parallel Acceleration:}
DoUT replaces iterative critique with parallel evaluation, significantly reducing the number of reasoning iterations. This approach enhances decision-making speed and accelerates robotic responses.

\textbf{4) Improved Spatial Awareness:}
The Evaluator compensates for the Proposer's limitations, enhancing overall performance in 3D environments.

\textbf{5) Flexible Output:}
Through a scoring mechanism, the Evaluator provides a ranked list of candidate propositions, allowing the robot to dynamically adjust its decisions in real-time based on the actual environmental conditions.

DoUT provides a streamlined and adaptive solution that addresses the limitations of DoT for real-time, responsive reasoning in robotic applications, enhancing efficiency, stability, and adaptability.

\subsubsection{Evaluating the Performance of DoT and DoUT}
To evaluate the performance of DoUT, a comparative experiment with DoT is conducted to measure the number of LLM reasoning iterations required as the number of tasks increases. Both methods use identical evaluation criteria, with the input consisting of a sequence of panoramic images captured during the exploration process. The comparison results are presented in Fig.~\ref{fig:compare_dot_dout}.

As shown in Fig.~\ref{fig:compare_dot_dout}, DoUT stabilizes after 3–4 task executions, while DoT requires approximately 10–11 task executions to achieve stability. This efficiency gain is primarily due to DoUT's structured feedback mechanism, which provides the LLM with targeted, task-specific evaluation feedback. This enables the reasoning process to quickly adapt to the evaluation criteria and converge more rapidly. On the other hand, DoT lacks structured feedback, relying entirely on its internal processes for refinement, resulting in slower convergence across tasks.

In addition to faster convergence, DoUT significantly reduces the number of reasoning iterations required within each task. Once converged, DoUT performs with just 1 LLM call per reasoning cycle, whereas DoT requires 4 calls. This improvement stems from the introduction of an external Evaluator, which centralizes the evaluation process and replaces DoT's sequential internal critiques with parallel assessments. By evaluating candidate propositions simultaneously, DoUT minimizes the total iteration count while maintaining high reasoning accuracy. In contrast, DoT's sequential evaluation process involves step-by-step critiques and validations, resulting in longer reasoning cycles as multiple internal steps are required to produce an output. This comparison highlights DoUT's superior efficiency and adaptability, making it particularly advantageous for real-time robotic applications where rapid and responsive decision-making is critical.

\subsection{DoUT Evaluation and Results}
\label{sec:dout_evaluation}
This section presents the evaluation of the DoUT across multiple state-of-the-art LLMs, including GPT-4o, GPT-4o-mini, LearnLM 1.5, and Moonshot AI. We assess the learning performance of these models using cosine similarity to compare the candidate propositions generated by the LLMs with the ranked propositions provided by the Evaluator. The results demonstrate how DoUT enhances reasoning accuracy and efficiency, enabling LLMs to adapt their outputs effectively over multiple iterations in real-time robotic applications.

\begin{figure}[tbp]
    \centering
    \includegraphics[scale=0.3]{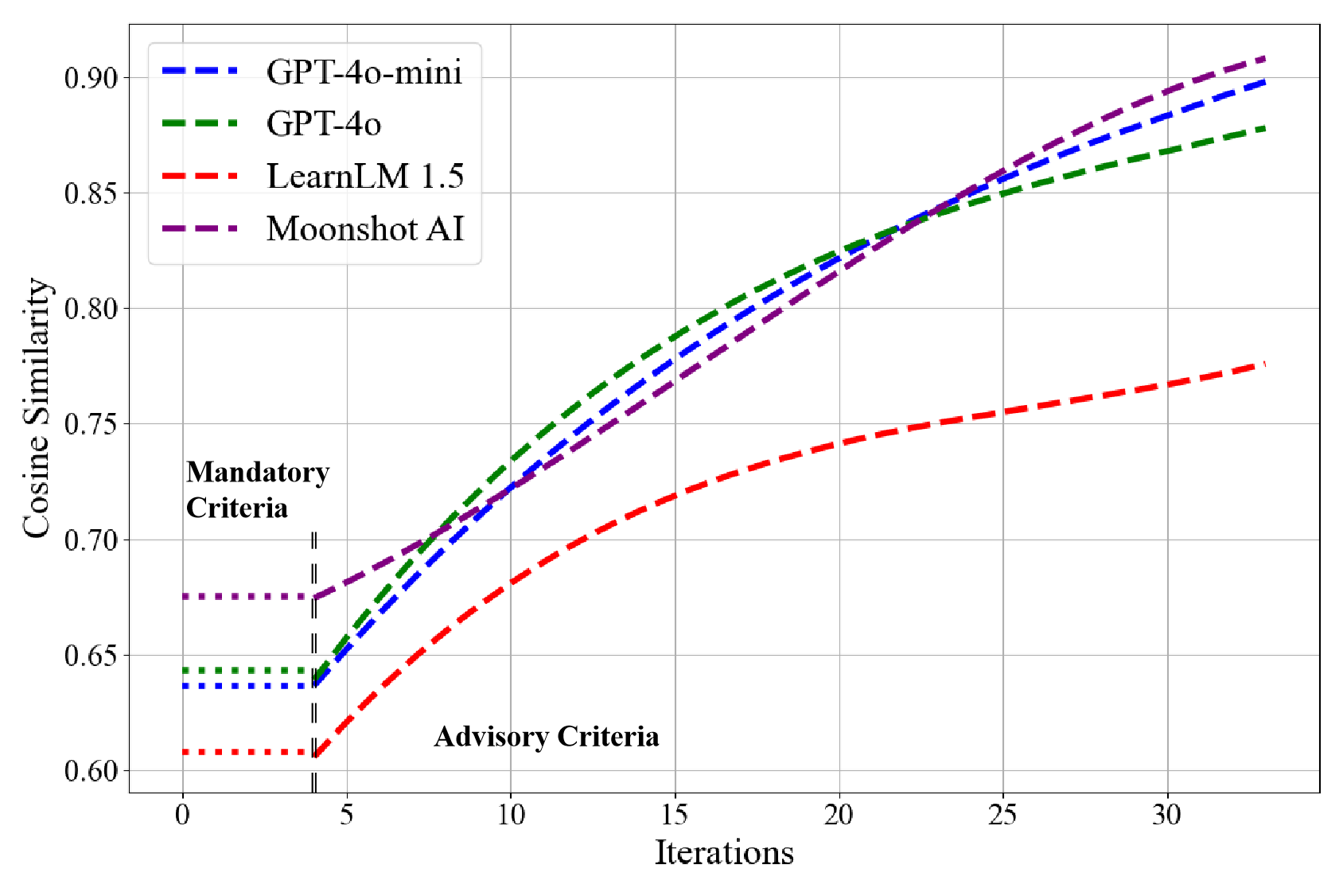}
    \caption{Comparison of LLMs' learning performance under DoUT, evaluated using cosine similarity between the candidate propositions $\mathcal{D}$ provided by the LLM and the ranked propositions $\mathcal{D}^*$ provided by the Evaluator. The initial straight lines represent the primary impact of the Mandatory Criteria, while the subsequent curves illustrate the primary effect of the Advisory Criteria.}    
    \label{fig:eval_dout}
\end{figure}

\subsubsection{Evaluation of DoUT across Multiple LLMs}
To analyze how DoUT performs across different LLMs, experiments are conducted using several state-of-the-art LLMs, including GPT-4o, GPT-4o-mini \cite{achiam2023gpt}, LearnLM 1.5 \cite{learnlm}, and Moonshot AI \cite{moonshot2023kimi}. These models support multimodal inputs (text and images) and demonstrate strong capabilities in image understanding and task reasoning. For a fair comparison, all models are evaluated using the same prompts and environmental panoramas. The cosine similarity metric is used to measure the alignment between the candidate propositions $\mathcal{D}$ generated by the LLMs and the ranked propositions $\mathcal{D}^*$ provided by the Evaluator. This metric quantifies the similarity between two vectors and is defined as:
\begin{equation}
    \text{cosine\_similarity}(\mathcal{D},\mathcal{D}^*) = \frac{\mathcal{D} \cdot \mathcal{D}^*}{||\mathcal{D}||||\mathcal{D}^*||}
\end{equation}

Fig. \ref{fig:eval_dout} illustrates the performance of DoUT across different LLMs, which can be divided into two stages: the Mandatory Criteria stage and the Advisory Criteria stage. During the Mandatory Criteria stage, the Evaluator provides feedback to the LLM, enabling it to adjust its reasoning to meet predefined requirements. As shown in Fig.~\ref{fig:eval_dout}, all LLMs fulfill the Mandatory Criteria within 1–4 iterations. However, due to the LLM's initial lack of familiarity with the Evaluator's Advisory Criteria and the distortions inherent in the equidistant cylindrical projection of panoramas, the candidate propositions $\mathcal{D}$ generated by the LLMs differ significantly from the ranked propositions $\mathcal{D}^*$ provided by the Evaluator. For example, at iteration 4, LearnLM 1.5 achieves a cosine similarity of only 0.61, whereas Moonshot AI demonstrates the best performance with a similarity of 0.67.

In the second stage, the Advisory Criteria are introduced. Feedback from the Evaluator allows the LLMs to infer and learn these criteria, leading to progressive alignment of their reasoning outputs with the Evaluator's rankings. As iterations increase, all LLMs show performance improvements to varying degrees, resulting in enhanced alignment between $\mathcal{D}$ and $\mathcal{D}^*$. For example, GPT-4o reaches a cosine similarity of 0.75, achieving this 12 iterations earlier than LearnLM 1.5. Moonshot AI attains the highest overall similarity of 0.91, whereas LearnLM 1.5 only reaches 0.78. Notably, GPT-4o-mini exhibits the largest improvement, increasing by 0.26, compared to LearnLM 1.5’s smaller gain of 0.17.

DoUT leverages the Mandatory and Advisory Criteria to provide structured feedback for LLM reasoning, refining even initially inaccurate outputs and thereby significantly enhancing the LLM's reasoning capabilities without task-specific pretraining. By incorporating parallelized evaluations, DoUT further improves efficiency, making its outputs more suitable for real-time robotic decision-making.

\begin{figure}[tbp]
    \centering
    \includegraphics[scale=1.2]{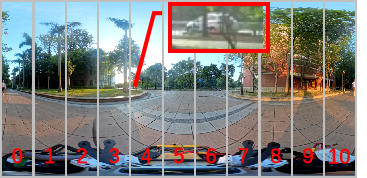}
    \caption{Comparison of different candidate propositions $\mathcal{D}$ after multiple iterations under the DoUT, using the same environmental panorama as input. Initially, before applying DoUT, the candidate propositions are $\mathcal{D} = \{4, 5, 6, 7, 3, 8, 9, 10, 2, 1, 0\}$. After 10 iterations, the sequence becomes $\mathcal{D} = \{4, 5, 6, 7, 3, 8, 9, 1, 2, 0, 10\}$. After 15 iterations, the sequence changes to $\mathcal{D} = \{4, 5, 6, 3, 7, 8, 9, 2, 1, 0, 10\}$. The ranked candidate propositions provided by the evaluator are $\mathcal{D}^* = \{4, 5, 6, 3, 7, 8, 2, 9, 1, 0, 10\}$.}    
    \label{fig:dout_details}
\end{figure}

\begin{figure*}[ht]
    \centering
    \subfigure[The final candidate propositions are $\{D4; D5; D3; D6; D7; D8; D1; D9; D2; D0; D10\}$. The LLM prioritizes D4 and D5, where vehicles are visible and potentially indicate a parking area.] {
        \includegraphics[width=1.0 \linewidth]{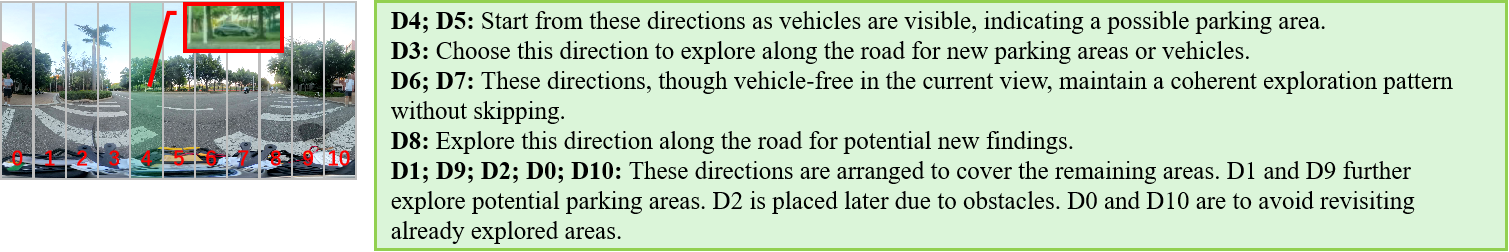}
    }
    \subfigure[The final candidate propositions are $\{D4; D5; D6; D7; D3; D2; D8; D9; D1; D0; D10\}$. The LLM prioritizes exploring the strategic junction D4, as the robot is near the scene boundary, and no areas likely to contain vehicles are visible in the current view.] {
        \includegraphics[width=1.0 \linewidth]{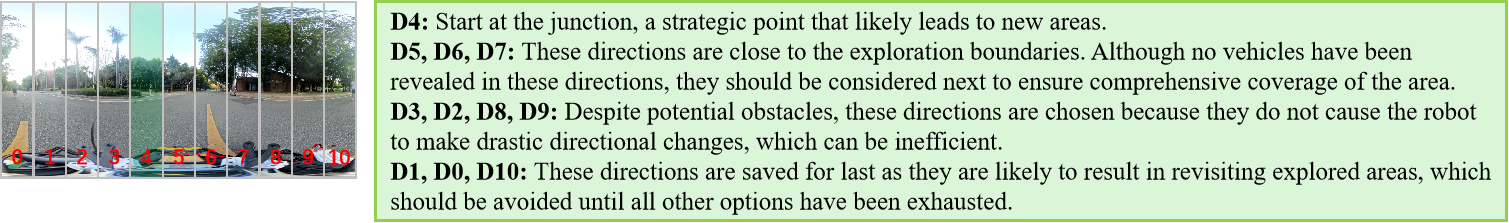}
    }
    \subfigure[The final candidate propositions are $\{D0; D10; D5; D4; D6; D1; D9; D3; D7; D2; D8\}$. The LLM prioritizes turning toward D0 and D10 to move closer to the target location in Building A, as the robot is currently positioned in Building D.] {
        \includegraphics[width=1.0 \linewidth]{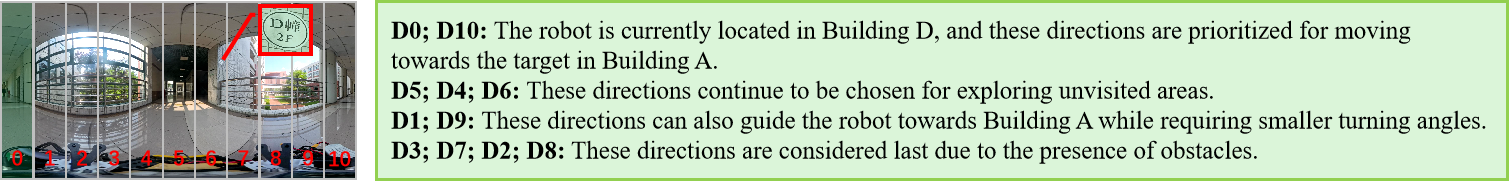}
    }
    \subfigure[The final candidate propositions are $\{D4; D5; D6; D3; D7; D2; D8; D1; D9; D0; D10\}$. After reaching Building A, D4 is prioritized for further exploration, as it offers a higher likelihood of locating the target.] {
        \includegraphics[width=1.0 \linewidth]{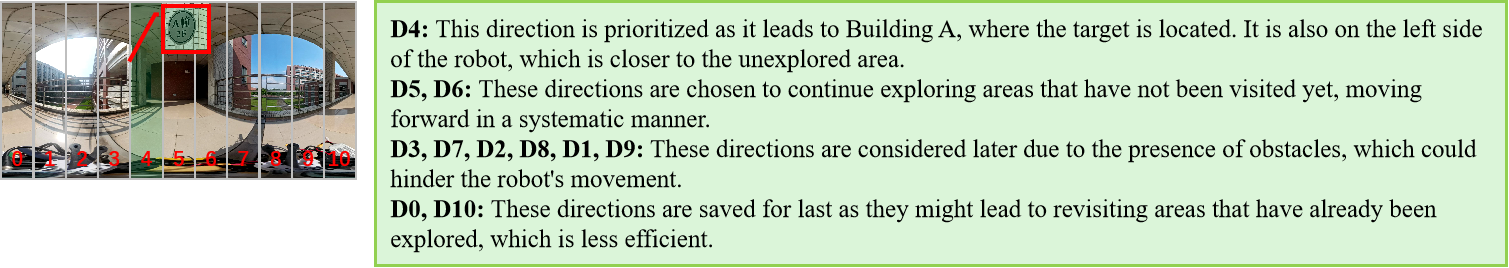}
    }
    \caption{The candidate propositions and their corresponding reasoning basis inferred by the LLM for different search tasks after multiple iterations of the DoUT. The panorama on the left represents the real-time environment, with red boxes highlighting the local image details relied upon by the LLM during reasoning. On the right, the green boxes present the LLM's reasoning logic for generating the candidate propositions. The LLM analyzed the environment from the input panorama to reason about potential target locations and generated corresponding candidate propositions. (a) and (b) illustrate the reasoning outputs for locating the target vehicle in Scene 1, while images (c) and (d) depict the reasoning outputs for searching for the target classroom in Scene 2.}
    \label{fig:search_process_details}
    % \vspace{-5mm}
\end{figure*}

\subsubsection{LLM Reasoning Insights under DoUT}

Fig.~\ref{fig:dout_details} illustrates the performance of DoUT during the search task, highlighting how the LLM's candidate propositions become increasingly aligned with the ranked candidate propositions provided by the Evaluator as the number of search task executions increases. Initially, without task-specific pretraining, the LLM, while capable of understanding the environmental panorama and reasoning about potential target areas, may not provide the most accurate recommendations. For example, the distortions present in the panorama can affect its distance perception, causing the robot to recommend previously explored areas. Additionally, the LLM may provide propositions that, without considering the robot's actual state, lead to unnecessary course corrections. Despite these initial challenges, as the robot continues to execute tasks and receives feedback, the LLM's performance improves. Over time, the LLM’s reasoning $\mathcal{D}$ gradually aligns with the ranked candidate propositions $\mathcal{D}^*$ provided by the Evaluator. This demonstrates that DoUT's feedback-driven process enables the LLM to learn and refine its reasoning accuracy, ultimately providing more accurate and efficient candidate propositions that guide the robot toward the most promising areas for target detection.

Figs.~\ref{fig:search_process_details} illustrate key decision-making steps during the robot's target search process. On the left of each subfigure is the environmental panorama, with the red-highlighted regions indicating the key areas relied upon by the LLM during reasoning. The green boxes on the right represent the main logic behind the LLM's candidate propositions.

In Scene 1, when the robot detects vehicles in the area, even if they are not the target vehicle, the LLM recommends exploring that area because it is likely a parking lot, as shown in Fig.~\ref{fig:search_process_details}(a). When the robot approaches the boundary of the search region and no potential target locations are visible within its current view, the LLM recommends exploring unexplored branches, as shown in Fig.~\ref{fig:search_process_details}(b). Similarly, in Scene 2, the LLM demonstrates effective reasoning during the search for the target classroom. For example, while the robot is exploring Building D, the LLM prioritizes turning around and moving toward Building A, where the target is located, as illustrated in Fig.~\ref{fig:search_process_details}(c). Upon reaching Building A, the LLM further recommends moving toward directions with a higher likelihood of locating the target, as shown in Fig.~\ref{fig:search_process_details}(d).

\end{document}